\documentclass[sigconf]{acmart}

\usepackage{booktabs} 
\usepackage{amsmath,graphicx}
\usepackage{multirow}
\usepackage{pifont}
\usepackage{subcaption}
\usepackage{graphicx}
\usepackage{tabularx}
\usepackage{xcolor}
\usepackage{colortbl}
\usepackage{pgfplots}
\usepackage{pgfplotstable}
\usepackage{enumerate}
\pgfplotsset{compat=1.7}
\usepackage{tikz}
\usepackage{makecell}
\usepackage{amsmath}
\usepackage{breqn}
\usepackage[utf8]{inputenc}
\usepackage{microtype}
\hypersetup{colorlinks = true,
            linkcolor = blue,
            urlcolor  = blue,
            citecolor = blue,
            anchorcolor = blue}

\newcommand{\datafull}{Wikipedia Reference Logo Dataset}
\newcommand{\data}{WiRLD}

\copyrightyear{2022}
\acmYear{2022}
\setcopyright{acmcopyright}\acmConference[ICVGIP'22]{Proceedings of the Thirteenth Indian Conference on Computer Vision, Graphics and Image Processing}{December 8--10, 2022}{Gandhinagar, India}
\acmBooktitle{Proceedings of the Thirteenth Indian Conference on Computer Vision, Graphics and Image Processing (ICVGIP'22), December 8--10, 2022, Gandhinagar, India}
\acmPrice{15.00}
\acmDOI{10.1145/3571600.3571625}
\acmISBN{978-1-4503-9822-0/22/12}

\editor{Soma Biswas}
\editor{Shanmuganathan Raman}
\editor{Amit K Roy-Chowdhury}

\acmArticle{25}

\begin{document}

\title{Contrastive Multi-View Textual-Visual Encoding: Towards One Hundred Thousand-Scale One-Shot Logo Identification}
\titlenote{Produces the permission block, and
  copyright information}
  
  \author{Nakul Sharma}
  \orcid{0000-0003-2218-4624}
  \affiliation{%
    \institution{Indian Institute of Technology}
    \streetaddress{}
    \city{Jodhpur}
    \state{Rajasthan}
    \country{India}
    \postcode{342030}
  }
  \email{sharma.86@iitj.ac.in}
  
   \author{Abhirama S. Penamakuri}
  \orcid{0000-0003-3646-8492}
  \affiliation{%
    \institution{Indian Institute of Technology}
    \streetaddress{}
    \city{Jodhpur}
    \state{Rajasthan}
    \country{India}
    \postcode{342030}
  }
  \email{penamakuri.1@iitj.ac.in}

    \author{Anand Mishra}
  \orcid{0000-0002-7806-2557}
  \affiliation{%
    \institution{Indian Institute of Technology}
    \streetaddress{}
    \city{Jodhpur}
    \state{Rajasthan}
    \country{India}
    \postcode{342030}
  }
  \email{mishra@iitj.ac.in}
\renewcommand{\shortauthors}{}

\begin{abstract}
In this paper, we study the problem of identifying logos of business brands in natural scenes in an open-set one-shot setting. This problem setup is significantly more challenging than traditionally-studied `closed-set' and `large-scale training samples per category' logo recognition settings. We propose a novel multi-view textual-visual encoding framework that encodes text appearing in the logos as well as the graphical design of the logos to learn robust contrastive representations. These representations are jointly learned for multiple views of logos over a batch and thereby they generalize well to unseen logos. We evaluate our proposed framework for cropped logo verification, cropped logo identification, and end-to-end logo identification in natural scene tasks; and compare it against state-of-the-art methods. Further, the literature lacks a `very-large-scale' collection of reference logo images that can facilitate the study of one-hundred thousand-scale logo identification. To fill this gap in the literature, we introduce \underline{Wi}kidata \underline{R}eference \underline{L}ogo \underline{D}ataset (WiRLD), containing logos for 100K business brands harvested from Wikidata. Our proposed framework that achieves an area under the ROC curve of 91.3\% on the QMUL-OpenLogo dataset for the verification task, outperforms state-of-the-art methods by 9.1\% and 2.6\% on the one-shot logo identification task on the Toplogos-10 and the FlickrLogos32 datasets, respectively. Further, we show that our method is more stable compared to other baselines even when the number of candidate logos is on a 100K scale.
\end{abstract}

%

\ccsdesc[500]{Computing methodologies~Image representations}

\keywords{supervised contrastive learning, one-shot learning, open-set recognition, logo identification.}

\maketitle

\section{Introduction}
\label{sec:intro}
\begin{figure}[!t]
\begin{minipage}[b]{1.0\linewidth}
  \centering
  \centerline{\includegraphics[width=8.25cm]{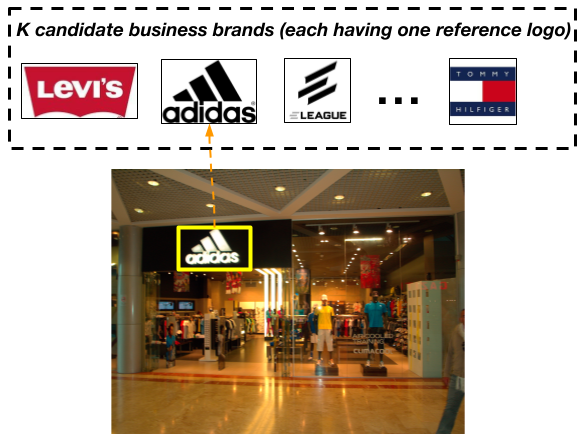}}
  \caption{\label{fig:goal} Our Goal: given a natural scene and a gallery of ``one'' reference logo each for $K$ ``unseen'' business brands, our goal is to identify the correct logo. We present a novel contrastive multi-view textual-visual encoding to address this problem. Further, for the first time in the literature, we study the problem of logo identification in an extremely challenging scenario when the number of candidate logos, i.e. $K$ is as large as 100K.}
\end{minipage}
\end{figure}

We study the problem of logo recognition in a practical setting where ``only one'' reference logo each for $K$ ``unseen'' business brands is available during inference, and the task is to detect the logo in a natural scene and identify it as one of the $K$ potential logos. We refer to this problem as \emph{Open-set One-shot Logo Identification in the Wild} and illustrate it in Figure~\ref{fig:goal}. The success of this challenging task can lead to many downstream real-world applications, including comprehensive scene understanding, and image search.

\emph{Open-set One-shot Logo Identification in the Wild} is a challenging task (especially when $K$ is of one-hundred-thousand scale) and requires a model to learn robust and discriminative encoding of logos that can generalize well even to unseen business brands. Inspired by the seminal works in contrastive multi-view encoding~\cite{tian2020contrastive, byol2020, chen2020simple, khosla2020supervised}, we present a supervised contrastive learning framework. Our framework encodes textual\footnote{Often business brand names are part of logos, our method leverages this fact while learning representation.} as well as visual features associated with the graphical design of logos and learns a fused robust representation using our novel supervised contrastive loss formulation. Our framework requires a set of cropped logos during training. During inference, our model, by virtue of these \emph{learned representations}, is able to compare unseen logos reasonably well even with an off-the-shelf method for detecting logos and na\"{\i}ve cosine similarity. Our framework differs from popular contrastive loss-based methods, e.g., pairwise~\cite{koch2015siamese} and triplet loss~\cite{hoffer2015deep} as it jointly optimizes the loss in a batch and learns a discriminative representation. 

Furthermore, there does not exist a dataset to study very large-scale logo identification in the literature. To fill this gap, we introduce \underline{Wi}kidata \underline{R}eference \underline{L}ogo \underline{D}ataset or \data{} in short -- a very-large-scale logo dataset containing reference logos for 100K business brands. We curate this dataset from an open-source knowledge base, namely Wikidata~\cite{wikidata} and use this curated set as a reference dataset in our very-large-scale logo identification experiment.
This collection can augment other datasets in the literature for performing large-scale logo identification experiments.

We perform rigorous experiments to evaluate our proposed model in three different settings: (i) cropped logo verification, (ii) cropped logo identification, (iii) end-to-end logo detection and identification,  and evaluate the performance of various relevant methods including ours over four public datasets, namely QMUL-OpenLogo~\cite{su2018open}, FlickrLogos-47~\cite{romberg2011scalable}, FlickrLogos-32~\cite{flickr27} and TopLogos~\cite{su2017deep}. Further, in order to perform truly very-large-scale logo identification, we use QMUL-OpenLogo dataset~\cite{su2018open}
as probe and our newly introduced dataset viz. \data{} as a reference set.  
Our method achieves area under the ROC curve of 91.3\% on the QMUL-OpenLogo dataset on cropped logo verification task. Further, our proposed framework outperforms state-of-the-art methods by 9.1\% and 2.6\% on the task of unseen cropped logo identification over TopLogos~\cite{su2017deep} and Flickr32~\cite{flickr27} datasets, respectively. 

\noindent\textbf{Contributions:}
To summarize, our contributions are three folds, (i) We present a contrastive multi-view encoding of visual-textual features by fusing textual, i.e., text associated with logos and visual, i.e., graphical design of logos and learn more robust and generalizable features. Our proposed contrastive multi-view encoding compels the samples from the same class and their augmented views closer and the samples from different classes and their augmented views farther in the semantic space. (ii) For the first time in the literature, we study the problem of logo identification in an extremely challenging scenario where the number of candidate logos is as large as 100K. In order to facilitate this study, we introduce a very-large-scale logo dataset, \datafull{} containing 100K reference logos. (iii) Our method achieves state-of-the-art results on the task of one-shot logo identification for unseen logos on four public logo datasets. Further, we also show the robustness of our approach for logo identification in a very-large-scale setting. We make our code and dataset available at our project website:  \url{https://vl2g.github.io/projects/logoIdent/}.

\section{Related Works}
\label{sec:relWork}

\subsection{Logo Recognition}
The majority of the successful logo recognition approaches, including traditional~\cite{joly2009logo, flickr27} as well as recent neural methods~\cite{romberg2013bundle, hoi2015logo, iandola2015deeplogo, bianco2015logo, bianco2017deep, bastan2019large, bhunia2019deep} pose the problem as a closed-set recognition problem, where all business brands are seen during training, and a large number of logos per business brand are available. This is not a  practical setting for real-world scenarios. Open-set logo recognition methods~\cite{fehervari2019scalable, li2022seetek} have been proposed to have a closer to a real-world setting but often relaxing one-shot assumptions. On one-shot logo recognition, recently~\cite{vargas2020one} reported the performance of the Siamese network. We experimentally compare against this approach and outperform it by a large margin. Slightly advanced one-shot learning methods like Variational Prototypical Encoder (VPE)~\cite{vpe} leverage prototype images by learning a mapping from real-world images to prototype images; representations learnt via this mapping aid the one-shot performance of the model. However, the following work, VPE++~\cite{vpe++} has shown that embeddings learnt by VPE suffer from the hubness problem and hence extended the VPE framework by proposing a multi-task loss formulation that reduces hubness. VPE++ method treats contrastive loss and classification loss as isolated losses as part of the multi-task loss. Through this work, we provide a supervised contrastive learning framework that jointly leverages classification and contrastive objectives. Our results show that our framework learns more generalizable representations, which are key for open-set one-shot identification tasks.

\begin{figure}[!t]
\begin{minipage}[b]{1.0\linewidth}
  \centering
  \centerline{\includegraphics[width=8cm]{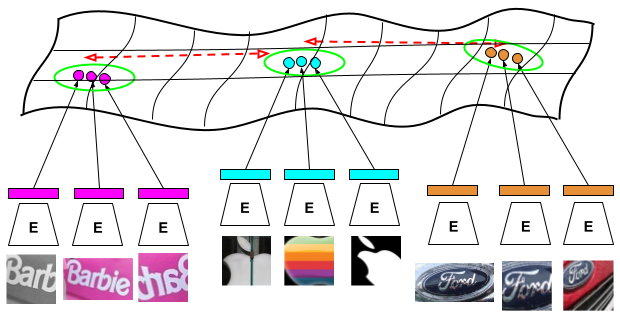}}
\caption{\label{fig:contra} Our proposed contrastive multi-view textual-visual encoding (E) (refer Section~\ref{sec:method} for more detail) projects logos in a subspace where multiple views of samples from the same and different business brands become closer and farther, respectively. We achieve this jointly for a batch using (2).}
\end{minipage}
\end{figure}

\subsection{Logo Datasets} Many logo datasets have been proposed for various tasks, including logo detection and classification. The majority of the existing datasets~\cite{flickr27, belgalogos, romberg2011scalable, su2018open, tuzko2017open, hoi2015logo, hou2017deep, logonet} have very limited coverage of logo classes, making them unsuitable for large-scale logo identification settings. Few works~\cite{logo2k, su2017weblogo, wang2022logodet, fehervari2019scalable, li2022seetek} have proposed logo datasets with more logo classes. However, unfortunately, only some of them are publicly available. Such limitations restrict existing models and benchmarks from exploring practical settings like very-large-scale logo identification tasks. To overcome such limitations and facilitate models to evaluate over the task of very-large-scale logo identification, we introduce a very-large-scale logo dataset, namely \datafull{} curated from open-source knowledge base Wikidata~\cite{wikidata}, containing 100K reference logos.

\subsection{Contrastive Learning}
Pairwise contrastive learning has been widely leveraged to learn generalizable features using Siamese networks~\cite{chopra2005learning, hadsell2006dimensionality}. Triplet loss uses triplets instead of pairs~\cite{hoffer2015deep}, where each triplet consists of an anchor, positive and negative samples, and the goal is to make the anchor closer to the positive sample and farther to the negative sample. However, the performance of these methods depends on the quality of pairs or triplets~\cite{vargas2020one}. Contrastive learning has been widely leveraged in the space of self-supervised representation learning approaches~\cite{jaiswal2021survey}. These methods rely on batch-wise losses~\cite{nce, npairloss} and their variants, where they do not sample negatives in isolation; instead, they use other batch samples as negatives. Authors in~\cite{khosla2020supervised} have extended contrastive learning to leverage class labels in loss formulation. In line with this research space, we present contrastive multi-view textual-visual encoding for robust and generalizable representation of logos.

\section{Proposed Approach}
\label{sec:method}
\subsection{Task Formulation}
In this work, we address \emph{open-set one-shot logo identification} in the following problem setup -- during training, images of cropped logos from a set of business brands ($Brand_{train}$) are available. However, during inference, given a natural scene and a set of $K$ business brands ($Brand_{test}$) with one reference logo for each brand, our goal is to localize and identify the logo in the scene. Here, it should be noted that  $Brand_{train} \cap Brand_{test}=\phi$ in our setup. In other words, we aim to identify unseen business brands during the inference. Learning discriminative and robust encoding for logos is required to address this task. To this end, we propose a \emph{contrastive multi-view textual-visual encoding} for addressing the problem. 

\subsection{Contrastive Multi-View Textual-Visual Encoding}
\subsubsection{Image representation}

\noindent For a given batch of $n$ logos $~\mathcal{I} = \{I^1, I^2,\cdots, I^n\}$ (where each $I^i \in \mathbb{R}^{3\times H \times W}$) sampled from a dataset, we begin by obtaining two distorted views of each image using a set data augmentations $\mathcal{A}$ adopted from ~\cite{pmlr-v139-zbontar21a}. The augmented views thus obtained, $\mathcal{I}_a$ and $ \mathcal{I}_b$ for each image in a batch are fed to the visual encoder $f_\theta$ and the textual encoder $g$ simultaneously. It should be noted that logos are often composed of graphical design and text, and the encoders $f_\theta$ and $g$ are designed to capture and encode these attributes of logos\footnote{If no text is detected in the logo, $g$ outputs a zero vector.}. For encoding the visual features of the logo, any visual encoder can be used in our framework. We use ResNet50~\cite{he2016deep} as our visual encoder to obtain 2048-dimensional features representing the graphical design of logos. These features ${\mathbf{V}_a}$ and ${\mathbf{V}_b}$, with $\mathbf{V}_{\{a,b\}} \in \mathbb{R}^{n\times{2048}}$, are obtained from both the views of logo $\mathcal{I}_a$ and $\mathcal{I}_b$ respectively.



\subsubsection{Text representation}

\noindent Any state-of-the-art scene text recognizer can be used to encode the textual features. We use the implementation from ~\cite{baek2019STRcomparisons} based on the CRNN ~\cite{Shi2017AnET} model (referred to as OCR-net in our framework). OCR-net has a traditional convolutional neural network to encode the image, followed by an LSTM module to decode the OCR-text character by character. We use the last hidden-state representation of the LSTM module as textual embedding. We refer to this module as our textual encoder $g$. We obtain the 256-dimensional textual feature vectors $\mathbf{T}_a$ and $\mathbf{T}_b$ for both the views of logo $\mathcal{I}_a$ and $\mathcal{I}_b$, respectively. Note that the weights of our textual encoder are frozen.

\subsubsection{Contrastive formulation and training objective}
\begin{figure*}[t!]
\centering
\includegraphics[width=17cm]{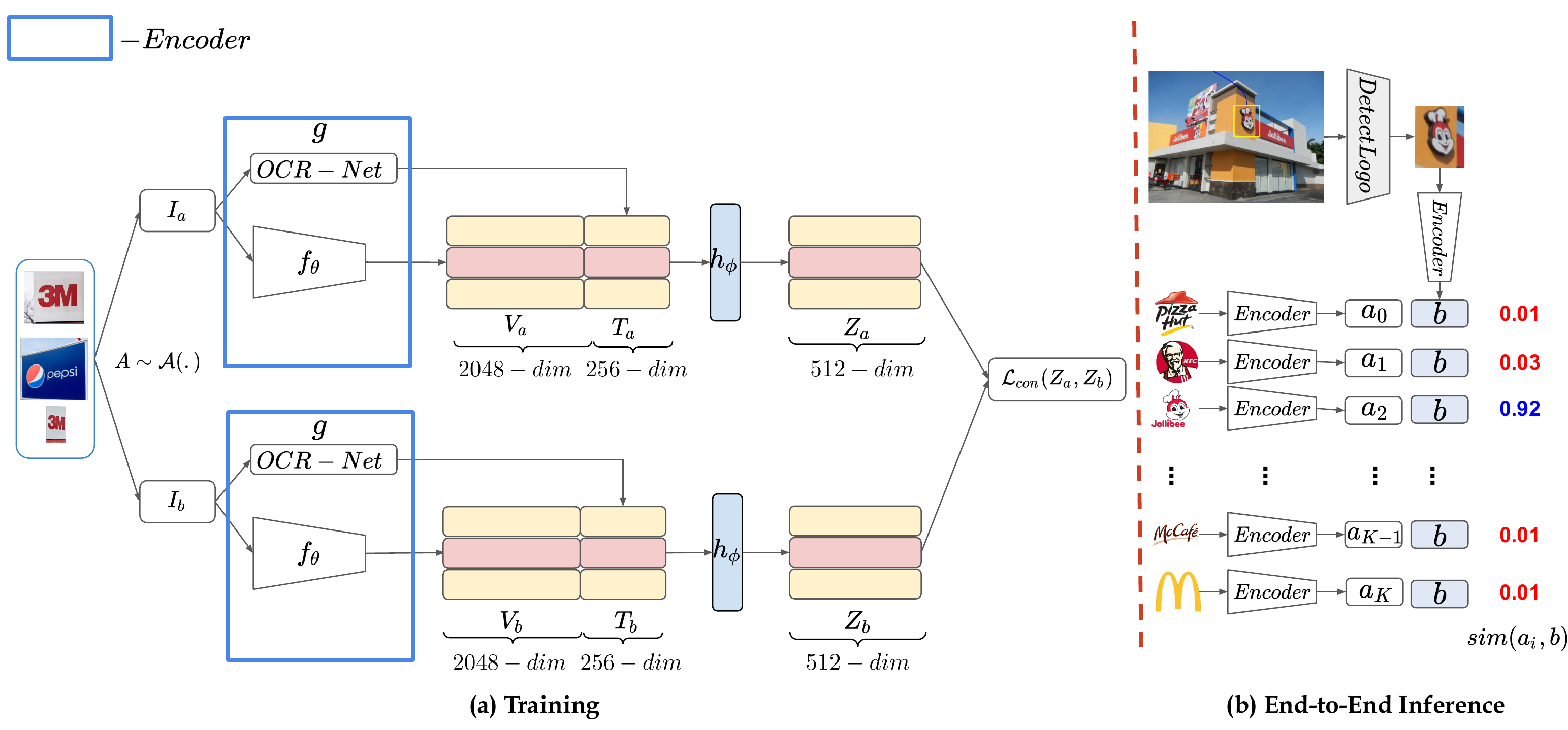}
\caption{\label{fig:method} \textbf{(a) Overview of our proposed framework.} We obtain two-views ${\mathcal{I}}_a$ and ${\mathcal{I}}_b$ of the input logo images for a batch $I$ using a set of data augmentations~\cite{pmlr-v139-zbontar21a}. For both-view batches $I_a$ and $I_b$, we obtain: textual embedding ($\mathbf{T}_a$, $\mathbf{T}_b$) obtained from $g({\mathcal{I}}_i)$, $\forall\ i \in {\{a,b\}}$, respectively; $g$ is last hidden-state vector of LSTM module of off-the-shelf OCR-Net~\cite{baek2019STRcomparisons}, and visual embeddings ($\mathbf{V}_a$, $\mathbf{V}_b$) obtained from $f_\theta(I_i)$, $\forall\ i \in {\{a,b\}}$, respectively; $f_\theta$ is visual backbone~\cite{he2016deep}. We concatenate ($\mathbf{V}_a:\mathbf{T}_a$), ($\mathbf{V}_b:\mathbf{T}_b$) and project using an MLP $h_\phi$ to obtain $\mathbf{Z}_{a}$ and $\mathbf{Z}_b$ respectively. The encoder is trained using the proposed supervised contrastive loss formulation. (b) illustrates the inference setup of our framework. Please refer to Section~\ref{sec:method} for more details. \textbf{[Best viewed in color].}}
\end{figure*}

Visual features $\mathbf{V}_a$ and $\mathbf{V}_b$ are then concatenated with textual features $\mathbf{T}_a$ and $\mathbf{T}_b$ respectively before being projected to a 512-dimensional space using an MLP $h_\phi$. The output embeddings  are normalized to obtain final logo representations $\mathbf{Z}_a$ and $\mathbf{Z}_b$, respectively, with $\mathbf{Z}_{\{a,b\}} \in \mathbb{R}^{n\times{512}}$, such that ${||v||_2 = 1}$ where {v} is any row vector in matrix $\mathbf{Z}_a$ and $\mathbf{Z}_b$. Parameters $\theta$ and $\phi$ are learnable. It should be noted here that each row of matrix $\mathbf{Z}_a$ and $\mathbf{Z}_b$ denote normalized feature vector corresponding to one image in a batch. 
An overview of our proposed framework is illustrated in Figure~\ref{fig:method}(a). (Notations used in our method are summarized in Table~\ref{tab:notations}.

Once we obtain $\mathbf{Z}_a$ and $\mathbf{Z}_b$, we formulate our contrastive loss function based on the intuition that the embeddings of the logos of the same brands across $\mathbf{Z}_a$ and $\mathbf{Z}_b$ should lie closer in the embedding space, while the embeddings of the logos of different categories should lie farther apart. Our objective is illustrated in Figure~\ref{fig:contra}. Formally, we define our loss function as follows:
\begin{dmath}
   \mathcal{L}_{con}({\mathbf{Z}_a}, \mathbf{Z}_b) = l(\mathbf{Z}_a, \mathbf{Z}_b) + l(\mathbf{Z}_b, \mathbf{Z}_a)\newline + l(\mathbf{Z}_a, \mathbf{Z}_a) + l(\mathbf{Z}_b, \mathbf{Z}_b),
\end{dmath}
where
 \begin{align}
  l(\mathbf{Z}_u, \mathbf{Z}_v) = -\sum_{i=1}^{i=n}\sum_{p\in{P(i)}}{\log\frac{exp(\mathbf{z}^{u}_i.\mathbf{z}^{v}_p/\tau)}{\sum\limits_{j=1}^{j=n}\mathop{}_{\mkern-5mu j\notin{P(i)}}{exp(\mathbf{z}^{u}_i.\mathbf{z}^{v}_j/\tau)}}}.
\end{align}   
Here, $i$ is an anchor in $\mathbf{Z}_u$, $P(i)$ is the set of all the positive logo indices corresponding to the anchor in the $\mathbf{Z}_v$ matrix. $\mathbf{z}_i^u$ is the $i^{th}$ row in $\mathbf{Z_u}$, similarly, $\mathbf{z}_i^v$ is the $i^{th}$ row in $\mathbf{Z}_v$. Parameter $\tau$ is empirically chosen as $0.07$ for all our experiments.

\begin{table}[!t]
    \centering
      \caption{\label{tab:notations}Notation used in the paper.}
    {
      \begin{tabular}{l r}
        \toprule
        \textbf{Symbol} &  \textbf{Meaning}\\
        \midrule
        $f_\theta$ & Visual Encoder\\
        $g$ & Textual Encoder\\
        $h_\phi$ & Projection MLP \\
        ${\mathcal{I}_{\{a, b\}}}$ & Augmented views of a batch\\
         ${\mathbf{V}_{\{a, b\}}}$ & Visual Features\\
       ${\mathbf{T}_{\{a, b\}}}$ & Textual Features \\
       ${\mathbf{Z}_{\{a, b\}}}$ & Projected final representation\\
        
       \bottomrule
      \end{tabular}}
\end{table}

Unlike other recently proposed supervised contrastive loss~\cite{khosla2020supervised}, for a given anchor, our loss formulation does not try to maximize the similarity scores for all the positive pairs over ``all the possible'' pairs in the batch. Instead, we maximize the cosine similarity of all the positive pairs over all the negative pairs only. This ensures that multiple positive pairs do not compete against each other to achieve a higher similarity score, thereby resulting in robust representations for logos, which is desirable for our task. Further, our proposed method is not only trained to learn the alignment between positive pairs in a batch but also learn to align different views of positive pairs; and similarly learns to push the embeddings of negative pairs as well as different views of negative pairs, farther from the positive pairs in the representation embedding space.

\subsection{Inference}
For end-to-end inference, given a natural scene, we detect logos using YOLOv5s~\cite{yolov5}, which is independently fine-tuned on the training set of QMUL-OpenLogo for the task of class-agnostic logo detection. 
Detected candidate logo bounding boxes are encoded using our ``trained'' contrastive multi-view textual-visual encoder that concatenates 2048-dimensional visual embedding from $f_\theta$ with 256-dimensional textual embedding from $g$ to obtain $\mathbf{b}$ to obtain a 2348-dimensional fused embedding. 
Reference logos for $K$ business brands (one reference logo per brand) are encoded in a similar fashion to obtain their corresponding fused embeddings $\{\mathbf{a}_1, \mathbf{a}_2, \cdots, \mathbf{a}_K\}$, with $\mathbf{a}_{\{1,\cdots , K\}} \in \mathbb{R}^{1\times2348}$. 
We rank the `K' reference logos based on the cosine similarity between $\mathbf{a}_i$ and $\mathbf{b}$ for $i = \{1,2,\cdots, K\}$ and take the most similar (= higher cosine similarity) as the identified logo. An overview of our inference setting is illustrated in Figure~\ref{fig:method}(b).

\subsection{Training and Implementation details} 
We use ResNet50~\cite{he2016deep} initialized with ImageNet pre-trained weights and frozen off-the-shelf OCR-Net~\cite{baek2019STRcomparisons}, and LSTM embeddings of the detected OCR-Text as our visual and textual backbones, respectively. We train our encoder with the proposed supervised contrastive loss framework using the SGD algorithm with a momentum of 0.9 and a learning rate of $1e-4$. We train all of our models on Nvidia GTX 1080 Ti GPU. During end-to-end inference, we utilize a class-agnostic YOLOv5s~\cite{yolov5} detector fine-tuned on our training split of the QMUL-OpenLogo dataset~\cite{su2018open} to detect logos from natural scene images. Additionally, we utilize a synthetic logo from each class in our formulation to have a better intra-class alignment during the experimental setting of \cite{vpe++}. We make implementation of this work available at our project website: \url{https://vl2g.github.io/projects/logoIdent/}.

\section{Experiments and Results}
\label{sec:expts}
In this section, we first discuss existing datasets that we use as part of our experimental settings in Section~\ref{sec:datasets} and then we present our curated dataset, namely \datafull{} in Section~\ref{sec:our_data}. 
We discuss baselines and ablations in Section~\ref{sec:baselines} and Section~\ref{sec:ablations}, respectively. Further, we briefly explain various evaluation settings; and discuss the quantitative and qualitative results in Section~\ref{sec:quant} and Section~\ref{sec:qual}, respectively.
\begin{figure}[t!]
\centering
\begin{tikzpicture}
\begin{axis}[
    grid,
	xlabel=FPR,
	ylabel=TPR,
    height=7cm,
    width=1\columnwidth,
    legend style={at={(.28,.2)},anchor=west}]
\addplot[color=green, mark=x] coordinates {
(0.0, 0.0)
(0.0, 0.01335559265442404)
(0.0, 0.02671118530884808)
(9.820288716488265e-05, 0.06628694883629578)
(0.0008838259844839438, 0.13709123048217617)
(0.0027496808406167143, 0.22665226357654916)
(0.01148973779829127, 0.34105862712363744)
(0.037906314445644705, 0.46607090248453303)
(0.10124717666699401, 0.5947166846705293)
(0.20514583128743985, 0.7232642639693607)
(0.3525483649219287, 0.8213689482470785)
(0.5272513011882549, 0.8880487086320338)
(0.689580673671806, 0.9375429637631346)
(0.8163606010016694, 0.9668074241382697)
(0.9053324167730531, 0.9839929293921241)
(0.9616026711185309, 0.9917509574781499)
(0.9860551900225867, 0.9965628989492291)
(0.9962682902877344, 0.998330550918197)
(0.999116174015516, 0.9994107826770107)
(0.9996071884513404, 0.9998035942256702)
(1.0, 1.0)
};
\addplot[color=yellow, mark=triangle] coordinates {
(0.0, 0.0)
(0.0007856230973190612, 0.09692624963173917)
(0.0015712461946381223, 0.1272709417656879)
(0.0020622606304625357, 0.14867917116763232)
(0.0027496808406167143, 0.1686143572621035)
(0.004222724148089954, 0.18933516645389376)
(0.004910144358244132, 0.21074339585583815)
(0.006088579004222724, 0.23185701659628793)
(0.007267013650201316, 0.2513011882549347)
(0.008445448296179908, 0.2733968378670333)
(0.009918491603653148, 0.29716193656093487)
(0.012766375331434744, 0.32740842580771873)
(0.014926838849062162, 0.35500343710105076)
(0.018756751448492585, 0.38966905627025433)
(0.021015417853284885, 0.42669154473141513)
(0.027202199744672494, 0.4646960620642247)
(0.03554944515368752, 0.512128056564863)
(0.04861042914661691, 0.5669252676028675)
(0.07168810763036433, 0.6447019542374546)
(0.13267210055975645, 0.749779043503879)
(1.0, 1.0)
};
\addplot[color=brown, mark=diamond*] coordinates {
(0.0, 0.0)
(0.02671118530884808, 0.2690041249263406)
(0.041048806834920945, 0.3740915340797486)
(0.05253854463321222, 0.43724219210371246)
(0.06343906510851419, 0.48566097033981537)
(0.07090248453304528, 0.5280887841288548)
(0.07748207797309241, 0.5595167943429582)
(0.08720416380241579, 0.5994892948340208)
(0.09545320632426593, 0.6230603024945983)
(0.10173819110281843, 0.6461402474955804)
(0.10880879897868997, 0.6647024160282852)
(0.11646862417755081, 0.6838538597525045)
(0.1256014926838849, 0.7010410528383422)
(0.1358145929490327, 0.7217638970732666)
(0.14632230187567513, 0.741209978393243)
(0.1568300108023176, 0.7591828717344333)
(0.17303348718452322, 0.7802003535651149)
(0.19100461553569675, 0.8025928108426635)
(0.21594814887557695, 0.8305833824395993)
(0.25974663655111463, 0.8671184443134944)
(1.0, 1.0)
};

\addplot[color=blue, mark=x] coordinates {
(0.0, 0.0)
(0.0, 0.022881272709417658)
(0.0, 0.0628498477855249)
(9.820288716488265e-05, 0.11705784150054012)
(0.0001964057743297653, 0.18020229794755965)
(0.0012766375331434744, 0.24933713051163703)
(0.0026514779534518315, 0.3231857016596288)
(0.004615535696749484, 0.4043012864578219)
(0.010802317588137092, 0.48404203083570657)
(0.022095649612098596, 0.564863006972405)
(0.04262005302955907, 0.6420504762840028)
(0.0831778454286556, 0.7220858293233821)
(0.16056172051458312, 0.8038888343317293)
(0.28616321319846805, 0.876166159285083)
(0.4650888736128842, 0.9310615732102524)
(0.6774035156633605, 0.9712265540606894)
(0.8620249435333399, 0.9918491603653148)
(0.9632721202003339, 0.998330550918197)
(0.993322203672788, 0.9999017971128351)
(0.9993125797898458, 1.0)
(1.0, 1.0)
};

\addplot[color=red, mark=triangle*] coordinates {
(0.0, 0.0)
(0.0, 0.015908867720710988)
(0.0, 0.042914661691053714)
(0.0, 0.08690955514092115)
(0.0, 0.14337621526072866)
(0.0002946086614946479, 0.21427869979377392)
(0.0010802317588137092, 0.29234999508985565)
(0.002847883727781597, 0.3741530000982029)
(0.006481390552882255, 0.46047333791613476)
(0.013944809977413337, 0.5474810959442208)
(0.026809388196012963, 0.632622999116174)
(0.05803790631444564, 0.7196307571442601)
(0.11666502995188059, 0.8011391534911126)
(0.22321516252577825, 0.8704703918295198)
(0.38633015810664834, 0.9307669645487577)
(0.5982519886084651, 0.9685750761072376)
(0.805852892075027, 0.9897868997348522)
(0.9366591377786507, 0.997446724933713)
(0.9871354217814003, 0.9999017971128351)
(0.998330550918197, 1.0)
(1.0, 1.0)
};
\legend{Pre-trained ResNet50~\cite{he2016deep}: 82.8, LitW~\cite{tuzko2017open}: 83.5, Siamese~\cite{vargas2020one}: 85.4, Ours (V): \textbf{89.6}, Ours (V+T): \textbf{91.3}}
\end{axis}
\end{tikzpicture}
\caption{\label{fig:roc} ROC curves for cropped logo verification task on the QMUL-OpenLogo dataset~\cite{su2018open}. The legends show the area under the ROC metric corresponding to each method.}
\end{figure}
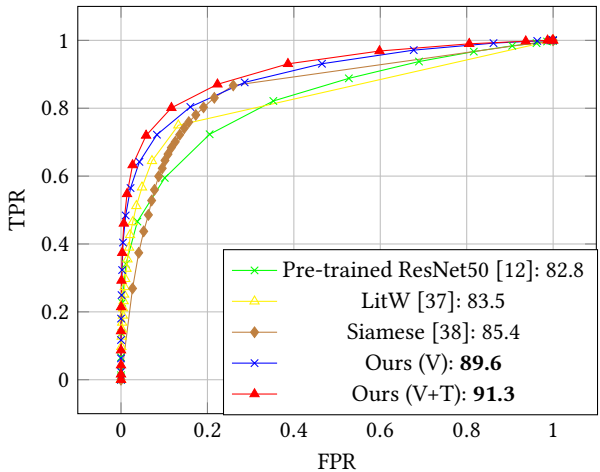

\begin{table}[!t]
    \centering
      \caption{\label{tab:datasetComp}Comparison of our newly introduced dataset, \datafull{} with the other related logo datasets.  Our introduced dataset provides a very-large-scale reference set for one-shot logo identification. ($^*$-not publicly available)}
    {
      \begin{tabular}{l r r}
        \toprule
        \textbf{Dataset} &  \textbf{\#logo classes} &  \textbf{\#images}  \\
        \midrule
        FlickrLogos-27~\cite{flickr27} & 27 &  1K \\
        FlickrLogos-32~\cite{flickr27} & 32 & 8.2K \\
        BelgaLogos~\cite{belgalogos}  & 37 & 10K \\
        FlickrLogos-47~\cite{flickr27} & 47 & 8.2K \\
       LOGO-Net~\cite{logonet} & 160 & 73.4K \\
       TopLogo-10~\cite{su2017deep} & 10 & 0.7K  \\
        Logo-405~\cite{hou2017deep} & 405 & 32.2K \\
        Logos in the wild~\cite{tuzko2017open} & 871 & 11K \\
        QMUL-OpenLogos~\cite{su2018open} & 300  & 27K \\
        WebLogo-2M~\cite{su2017weblogo} & 194  & 1.8M \\
        PL2K$^*$~\cite{fehervari2019scalable} & 2K  & 295K \\
        Logo-2K+~\cite{logo2k} & 2.3K  & 167K \\
        LogoDet-3K~\cite{wang2022logodet} & 3K & 158K \\
        PL8K$^*$~\cite{li2022seetek} & 8K  & 3M \\
        \makecell[l]{\textbf{\data{}} \textbf{(This work)}} & \textbf{100K} & \textbf{100K} \\
       \bottomrule
      \end{tabular}}
\end{table}

\subsection{Datasets}
\label{sec:datasets}
\begin{figure*}[!]
\begin{minipage}[b]{1.0\linewidth}
  \centering
  \centerline{\includegraphics[width=\textwidth]{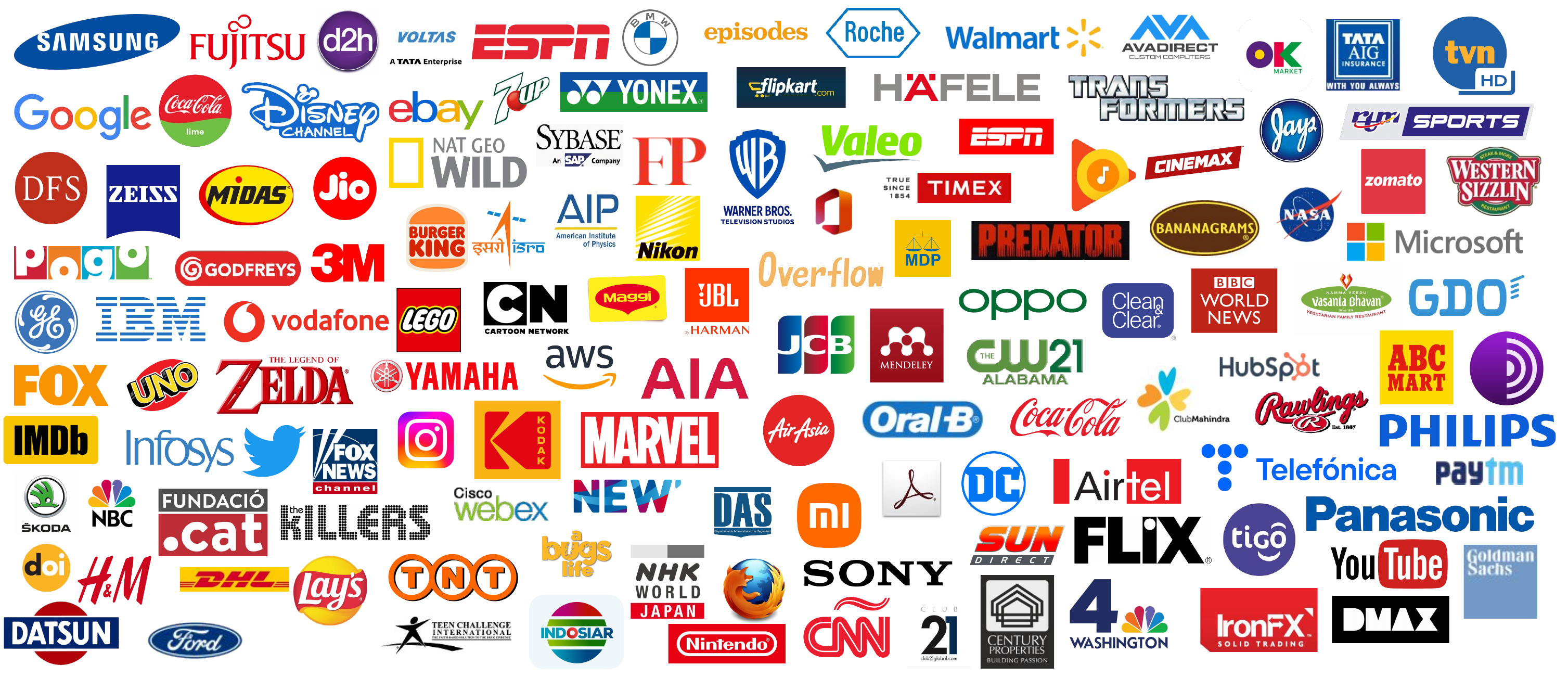}}
\caption{\label{fig:selectedwiki} A selection of logos from our newly introduced \datafull{}. In total, our dataset has around 100K logo classes, with each class having one reference logo. Note that these logos are noise-free and clean as they are sourced directly from Wikidata. Hence, it has great utility as a reference gallery set, especially for a task like very-large-scale one-shot logo identification.}
\end{minipage}
\end{figure*}

\subsubsection{\textbf{QMUL-OpenLogo Dataset ~\cite{su2018open}}} This dataset has 27K curated images of 336 business brands. We follow the same split as authors of~\cite{vargas2020one}, where logos from 211 business brands are used for training and fine-tuning, and one logo each from 125 business brands is used for testing. Note that train and test classes are disjoint. 

\subsubsection{\textbf{FlickrLogos-47 ~\cite{romberg2011scalable}}} It contains 2,235 annotated scenic images with logo regions spanning across 47 logo classes (32 symbolic logos and 15 textual logos). We randomly pick 30 business brands out of 47 for training and 17 unseen brands for testing purposes. We leverage the existing bounding box annotation for this dataset and thus obtain 1936 cropped logo images as part of the train set and 4032 as the test set. 

\subsubsection{\textbf{BelgaLogos~\cite{belgalogos}}} This dataset contains 10K logo images spanning over 26 logo classes. Following the setting in~\cite{vpe}, we use this dataset to train our model with our proposed framework.


\subsubsection{\textbf{Toplogos~\cite{su2017deep}}} This dataset consists of 700 logo images over ten logo classes. Following the setting in~\cite{vpe}, we use this dataset as a test dataset for our cropped logo verification task for a fair comparison with the baselines.

\subsubsection{\textbf{\datafull{}} \textbf{}{(\data{})}, \textit{(newly introduced in our work)}} 
\label{sec:our_data}
Many datasets have been proposed in the research space of logo detection, and recognition ~\cite{flickr27, belgalogos, romberg2011scalable, su2018open, tuzko2017open, hoi2015logo, hou2017deep, logonet, su2017weblogo, logo2k, wang2022logodet, fehervari2019scalable, li2022seetek}; however, unfortunately, the majority of these datasets have very limited coverage of logo classes or not publicly available; making them unsuitable for the tasks that demand a very-large-scale logo dataset, e.g. large-scale logo identification. (An overview comparing the various logo datasets is shown in Table~\ref{tab:datasetComp}). To overcome shortcomings of existing datasets and to facilitate models to explore the task of very-large-scale logo identification, we curate large-scale logos from an open-source knowledge base, namely Wikidata~\cite{wikidata}.
We follow a three-stage process to extract logos from Wikidata. In stage-1, we obtain all the entities over Wikidata with a logo with the help of the Wikidata SPARQL\footnote{\url{https://query.wikidata.org/}} query service. Once all entities are obtained, in stage-2, we parse the one-hop neighbourhood for each entity over the Wikidata graph and obtain logo URLs. Finally, in stage-3, we download original logo images from these URLs. We use this curated set of reference logo gallery for our task \textit{viz.} large-scale open-set one-shot logo identification. Our curated dataset has 100K reference logo images spanning over 100K logo classes (One logo image for every entity). The URLs of logo images of \data{} are available for download in our project website\footnote{\url{https://vl2g.github.io/projects/logoIdent/}}.


\begin{table}[!t]
    \centering
      \caption{\label{tab:cropped_logo_results_vpe}Comparison of \textbf{cropped logo identification} results on Flickr32~\cite{flickr27} and TopLogos-10~\cite{su2017deep} datasets, respectively. We report Top-1 accuracy for both seen and unseen logo classes. Baseline results for methods QuadNet~\cite{quadnet}, MatchNet~\cite{matchnet}, VPE~\cite{vpe} and VPE++~\cite{vpe++} are taken directly from~\cite{vpe++}.
     }
    
      \resizebox{1\columnwidth}{!}{
      \begin{tabular}{l r r r r}
        \toprule
        \multicolumn{1}{c}{} &   \multicolumn{2}{c}{Belga~\cite{belgalogos} $\xrightarrow{}$ Flickr-32~\cite{romberg2011scalable}} & \multicolumn{2}{c}{Belga~\cite{belgalogos} $\xrightarrow{}$ Toplogos~\cite{su2017deep}}
        \\
        \cmidrule(r){2-3}
        \cmidrule(r){4-5}
        Method & \makecell[c]{All \\ (Top-1)} & \makecell[r]{Unseen \\(Top-1)} & \makecell[c]{All \\ (Top-1)} & \makecell[r]{Unseen \\(Top-1)} \\
        \midrule
        VAE & 27.17 & 27.31 & 23.30 & 18.59\\
        Siamese Network~\cite{koch2015siamese} & 24.7 & 22.82 &  30.84 & 30.46 \\
        Pretrained ResNet~\cite{he2016deep} & 43.21 & 44.68 & 38.35 & 46.56 \\
        LitW~\cite{tuzko2017open} & 33.96 & 26.34 & 57.21 & 51.10 \\
        QuadNet~\cite{quadnet} & 31.68 & 28.55 & 38.89 & 34.16 \\
        MatchNet~\cite{matchnet} & 38.54 & 35.28 & 28.46 & 27.46 \\
        VPE~\cite{vpe} & 56.6 & 53.53 & 58.65 & 57.75 \\
        VPE++~\cite{vpe++} & 65.54 & 62.56 & 65.57 & 70.27 \\ 
        SupCon~\cite{khosla2020supervised} & 65.84 & 64.84 & 66.06 & 70.22\\
        \textbf{Ours - Vision} &  \textbf{66.42} & \textbf{64.92} & \textbf{72.05} & \textbf{72.49}
\\
        \textbf{Ours - Vision + Text} & \textbf{66.77} & \textbf{65.17} & \textbf{72.26} & \textbf{79.33} \\
        \bottomrule
      \end{tabular}}
\end{table}

\subsection{Baselines}
\label{sec:baselines}
We choose various state-of-the-art methods as baselines that are closely related to our problem setup. We group baselines into two categories, namely (i) single-stream methods and (ii) contrastive-loss based approaches. Under single stream networks, we use a pretrained ResNet~\cite{he2016deep} model and a method mentioned in LitW ~\cite{tuzko2017open}. Under contrastive-loss based approaches, we use the two approaches Siamese network-based approach~\cite{vargas2020one} and the recently proposed supervised contrastive loss-based approach~\cite{khosla2020supervised}. Additionally, we consider recent works, namely VPE++~\cite{vpe++}, VPE~\cite{vpe}, matching network~\cite{matchnet}, quadruplet networks~\cite{quadnet} and variational autoencoder as our baselines. For fair comparison against these additional baselines, we follow a similar experimental setup as~\cite{vpe++}. 

\begin{figure*}[!t]
\begin{minipage}[b]{1.0\linewidth}
  \centering
  \centerline{\includegraphics[width=\columnwidth]{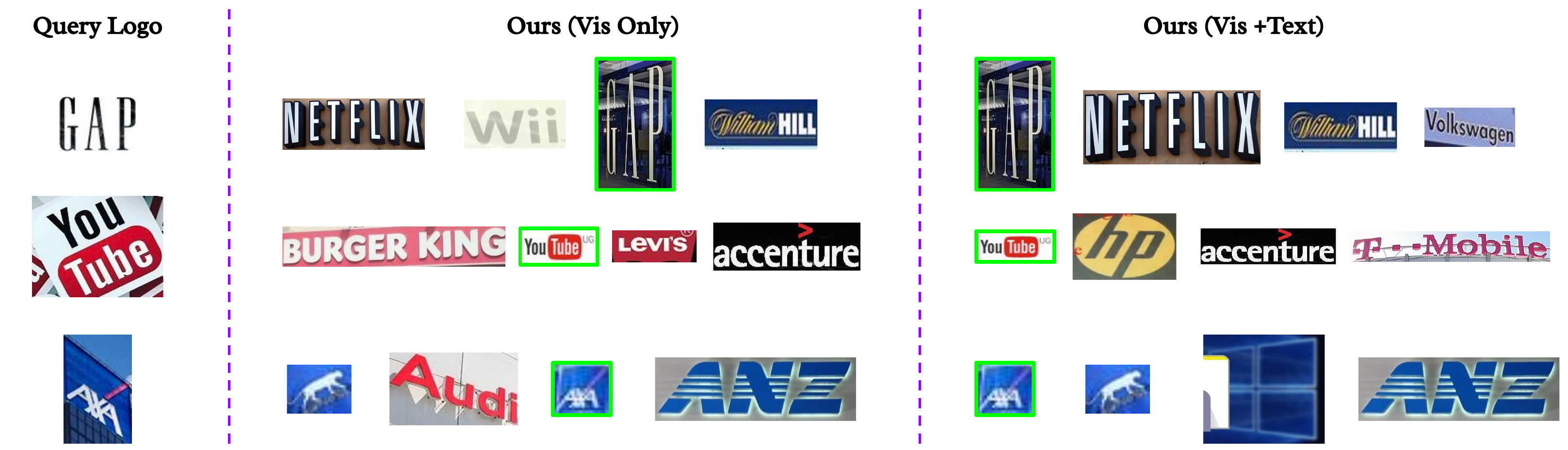}}
  \caption{\label{fig:visRes} A selection of test logos detected from the natural scene as queries. Each row has a query (on the left), and top-4 most similar logos obtained using Ours (Vision only) and Ours (Vision+Text) models on the cropped logo identification on the QMUL-OpenLogo. Logos with a green bounding box represent the correct match. These results show that our framework is able to learn robust representations leveraging both textual and visual cues from logos. [Best viewed in color].}
\end{minipage}
\end{figure*}

\begin{table}[!t]
    \centering
      \caption{\label{tab:cropped_logo_results}Comparison of \textbf{cropped logo identification} results on both QMUL-OpenLogo~\cite{su2018open} and FlickrLogos-47~\cite{romberg2011scalable} datasets.  We report Top-$k$ ($k=$ 1, 5 and 10) accuracy (in \%). 
    }
    \resizebox{1\columnwidth}{!}{
      \begin{tabular}{l r r r r r r}
        \toprule
        \multicolumn{1}{c}{} &   \multicolumn{3}{c}{QMUL-OpenLogo~\cite{su2018open}} & \multicolumn{3}{c}{FlickrLogos-47~\cite{romberg2011scalable}} \\
        \cmidrule(r){2-4}
        \cmidrule(r){5-7}
        Method & Top-1 & Top-5 & Top-10 & Top-1 & Top-5 & Top-10\\
        \midrule
        Levenshtein Distance & 30.8 & 34.1 & 34.1 & 17.6 & 17.6 & 29.4 \\
        Siamese Network~\cite{vargas2020one} & 23.3 & 49.2 & 61.7 & 41.2 & \textbf{94.1} & 94.1 \\
        Pretrained ResNet~\cite{he2016deep} & 30 & 48.3 & 59.2 & 29.4 & 82.4 & 88.2 \\
        LitW~\cite{tuzko2017open} & 27.5 & 54.2 & 68.3 & 17.6 & 76.5 & \textbf{100}\\
        SupCon~\cite{khosla2020supervised} & 44.2 & 62.5 & 70.8 & 76.5 & 88.2 & \textbf{100}\\
        \textbf{Ours - Vision} & 48.3 & 63.3 & 70 & 76.5 & \textbf{94.1} & 94.1 \\
        \textbf{Ours - Vision + Text} & \textbf{55.7} & \textbf{68.3} & \textbf{73.3} & \textbf{82.4} & \textbf{94.1} & 94.1\\
        \bottomrule
      \end{tabular}}
\end{table}

\begin{table}[!t]
    \centering
    \caption{\label{tab:end_to_end}Comparison of \textbf{end-to-end logo identification} results on both QMUL-OpenLogo~\cite{su2018open} and FlickrLogos-47~\cite{romberg2011scalable} datasets. We report Top-$k$ ($k=$ 1, 5 and 10) accuracy (in \%).
    }
    \resizebox{1\columnwidth}{!}{
      \begin{tabular}{l r r r r r r}
        \toprule
        \multicolumn{1}{c}{} & \multicolumn{3}{c}{QMUL-OpenLogo~\cite{su2018open}} & \multicolumn{3}{c}{FlickrLogos-47~\cite{romberg2011scalable}} \\
        \cmidrule(r){2-4}
        \cmidrule(r){5-7}
        Method & Top-1 & Top-5 & Top-10 & Top-1 & Top-5 & Top-10\\
        \midrule
        
        Levenshtein Distance & 16.6 & 19.2 & 22.5 & 0 & 5.9 & 17.6 \\
        Siamese Network~\cite{vargas2020one} & 12.9 & 25.9 & 39.7 & 43.8 & 81.2 & 87.5 \\
        Pretrained ResNet~\cite{he2016deep} & 16.4 & 28.4 & 39.7 & 43.8 & \textbf{87.5} & \textbf{93.8}\\
        
        LitW~\cite{tuzko2017open} & 17.2 & 33.6 & 43.1 & 43.8 & 81.2 & 87.5\\
        SupCon~\cite{khosla2020supervised} & 23.3 & 30.2 & 37.9 & \textbf{62.5} & 81.2 & \textbf{93.8}\\
        \textbf{Ours - Vision} & 24.1 & 32.8 & 41.4 & 56.2 & \textbf{87.5} & \textbf{93.8} \\
        \textbf{Ours - Vision + Text} & \textbf{26.7} & \textbf{39.7} & \textbf{48.3} & 56.2 & 81.2 & \textbf{93.8}\\
        \bottomrule
      \end{tabular}}
\end{table}
\subsection{Ablations}  
\label{sec:ablations}
We perform the following ablations, (i) our method's performance on seen classes: to benchmark and contrast the performance of our proposed framework over seen vs unseen logo classes, (ii) our method (without Text): to estimate the importance of textual pipeline, (iii) our method using different visual backbones: to estimate the role and importance of visual backbone. Further, to illustrate the performance of a method that only ranks the logos based on the recognized text and does not use visual cues, we also show results using Levenshtein distance between text detected from the logo and the reference logo crops. 

\subsection{Quantitative Results}
\label{sec:quant}
We quantitatively evaluate our proposed framework in four experimental settings and compare it with various related approaches. Note that the test set's classes (business brands) in all evaluation settings are unseen during training.

\subsubsection{\textbf{Cropped logo verification}} In this setting, a pair of cropped logos (from 20,000 logo image pairs~\cite{su2018open}) are compared against each other for a match. We present the ROC curve comparison of our framework with the baselines in Figure~\ref{fig:roc} on the QMUL-OpenLogo dataset. Our framework outperforms the previous state-of-the-art model by achieving an area under the ROC curve of 91.2\% on the QMUL-OpenLogo dataset.

\subsubsection {\textbf{Cropped logo identification}} In this task, we follow two settings: (i) Similar to ~\cite{vpe, vpe++} where a noise-free clean logo is matched over a set of cropped logos from natural scene images. We follow the same training and evaluation protocols, and we train our proposed framework on Belgalogo~\cite{belgalogos} dataset and evaluate over Flickr32~\cite{flickr27} and TopLogos-10~\cite{su2017deep} datasets, respectively, and baseline results are taken directly from~\cite{vpe++, vpe} for this setting; We present accuracy of seen vs unseen classes in Table~\ref{tab:cropped_logo_results_vpe}. Our framework outperforms the baselines on both seen and unseen categories. We have not included these baselines in further evaluation settings due to different training paradigms. (ii) Challenging setting where a noisy cropped logo is compared against `one' reference logo of $K$ business brands (where $K$ can be potentially large, and reference logos can be noisy as well). The reference logos are ranked based on similarity with the cropped logo. We compare Top-$k$ ($k=1$, 5 and 10) accuracy of our framework with the baselines in Table~\ref{tab:cropped_logo_results} on both QMUL-OpenLogo and FlickrLogos-47 datasets. On the QMUL-OpenLogo dataset, our vision-only encoder trained with the proposed loss framework has outperformed the baselines, indicating the robustness of the proposed loss formulation.

\begin{table}[!t]
    \centering
    \caption{\label{tab:visual_backbone}Logo identification results with our method over vision backbones, on QMUL-OpenLogo dataset~\cite{su2018open}. We report Top-$k$ ($k=$ 1, 5 and 10) accuracy (in \%).} 
    \resizebox{1\columnwidth}{!}{
      \begin{tabular}{l r r r r}
        \toprule
        \multicolumn{1}{c}{} & \multicolumn{1}{c}{} & \multicolumn{3}{c}{QMUL-OpenLogo~\cite{su2018open}} \\
        \cmidrule(r){3-5}
        Method & Vision backbone & Top-1 & Top-5 & Top-10\\
        \midrule
        Ours - Vision & AlexNet~\cite{krizhevsky2012imagenet} & 33.3 & 54.2 & 64.2 \\
        Ours - Vision + Text  & AlexNet~\cite{krizhevsky2012imagenet} & 35.0 & 52.5  & 66.7\\
        Ours - Vision & ResNet~\cite{he2016deep} & 48.3 & 63.3 & 70.0 \\
        Ours - Vision + Text  & ResNet~\cite{he2016deep} & \textbf{55.8} & \textbf{68.3} & \textbf{73.3}\\
        \bottomrule
      \end{tabular}}
\end{table}

\subsubsection{\textbf{End-to-end logo detection and identification}} This is the practical setting where we do not assume that cropped logos are provided to us. Instead, we first detect the logo and then compare it against reference logos. We compare Top-$k$ ($k=1$, 5 and 10) accuracy of our framework with the baselines in this setting as shown in Table~\ref{tab:end_to_end} on both QMUL-OpenLogo and FlickrLogos-47 datasets. On FlickrLogs-47, our method Top-1 accuracy is slightly inferior to one of the recent approaches. However, our Top-5 and Top-10 accuracy on this dataset are comparable.

\begin{figure*}[!t]
\begin{minipage}[b]{1.0\linewidth}
  \centering
  \centerline{\includegraphics[width=\columnwidth]{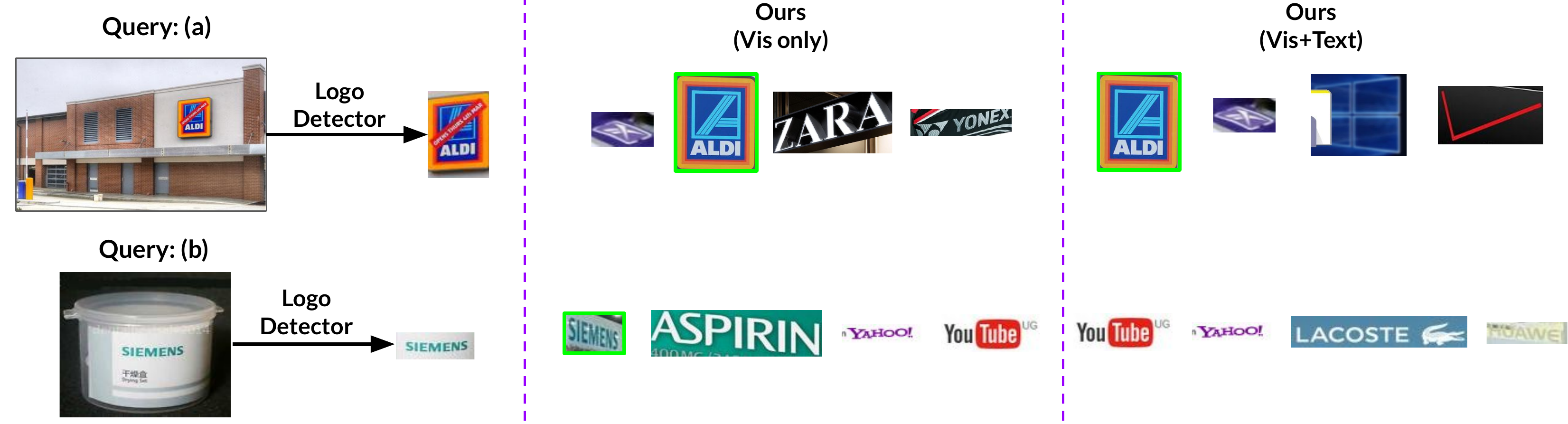}}
  \caption{\label{fig:e2evisRes} Logo identification from natural scene images. Each row has a natural scene query image (on the left), and top-4 most similar logos obtained using our proposed method over vision only and vision+text variants on the end-to-end logo identification setting on the QMUL-OpenLogo dataset~\cite{su2018open}. Logos with a green bounding box represent the correct match.}
\end{minipage}
\end{figure*}

\subsubsection{\textbf{Cropped logo identification against large-scale reference logos}} 
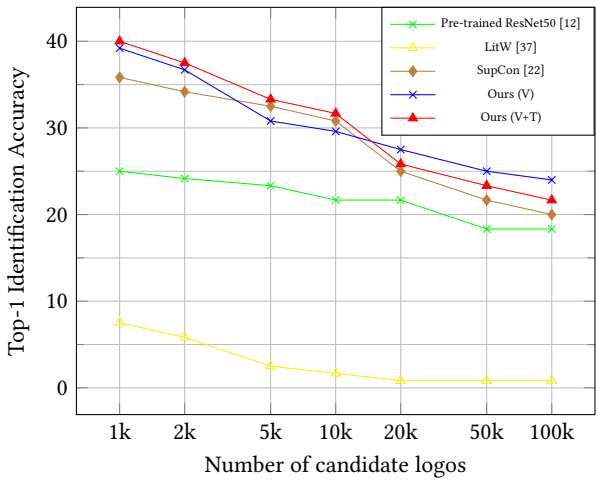
\begin{figure}[t!]
\begin{minipage}[b]{1.0\linewidth}
\centering
\begin{tikzpicture}
\begin{axis}[
    domain=0:100000,
	xlabel=Number of candidate logos,
	ylabel=Top-1 Identification Accuracy,
    height=7cm,
    xmode=log,
    xtick=data,
    xticklabels={1k, 2k, 5k, 10k, 20k, 50k, 100k },
    width=1\columnwidth,
    legend style={at={(.588,.843)},anchor=west},
    grid=both,
    minor tick num=1,
    ]
\addplot[color=green, mark=x] coordinates {
(1000, 25.0)
(2000, 24.16)
(5000, 23.33)
(10000, 21.67)
(20000, 21.67)
(50000, 18.33)
(100000, 18.33)
};
\addplot[color=yellow, mark=triangle] coordinates {
(1000, 7.5)
(2000, 5.83)
(5000, 2.5)
(10000, 1.67)
(20000, 0.83)
(50000, 0.83)
(100000, 0.83)
};
\addplot[color=brown, mark=diamond*] coordinates {
(1000, 35.83)
(2000, 34.17)
(5000, 32.5)
(10000, 30.8)
(20000, 25.0)
(50000, 21.67)
(100000, 20)
};

\addplot[color=blue, mark=x] coordinates {
(1000, 39.2)
(2000, 36.7)
(5000, 30.8)
(10000, 29.6)
(20000, 27.5)
(50000, 25)
(100000, 24)
};
\addplot[color=red, mark=triangle*] coordinates {
(1000, 40)
(2000, 37.5)
(5000, 33.3)
(10000, 31.67)
(20000, 25.83)
(50000, 23.33)
(100000, 21.66)
};
\legend{\tiny{Pre-trained ResNet50~\cite{he2016deep}}, \tiny{LitW~\cite{tuzko2017open}}, \tiny{SupCon~\cite{khosla2020supervised}}, \tiny{Ours (V)}, \tiny{Ours (V+T)}}
\end{axis}
\end{tikzpicture}
\caption{\label{fig:largescaleExp} Large-scale logo identification. We present Top-1 accuracy of our framework with the baselines over varying scales. Performance drop is expected with an increase of scale; however, our framework retains its performance over baselines owing to the robustness of learnt representations.}
\end{minipage}
\end{figure}

This setting enables us to evaluate the performance of our framework in real-world scenarios where a cropped logo is compared against a very large set of logo images with the scale ranging from 1K to 100K. We evaluate our proposed framework on the task of logo identification over the QMUL-OpenLogo dataset as a probe set along with our curated large-scale open-set one-shot \data{} as a reference set. Similar to the previous evaluation setting, we present Top-1 accuracy of our framework with the baselines over various scales of images in the gallery in a line chart in Figure~\ref{fig:largescaleExp}. In a large-scale logo identification setting, a performance drop is expected with an increase in scale. However, our results reported in Figure~\ref{fig:largescaleExp} suggest that the representations learnt by our framework remain robust when compared against the previous best-performing baseline SupCon~\cite{khosla2020supervised}. Our vision-only method slightly outperforms the vision-text method at higher scales, owing to the training constraints of OCR-Net, e.g. indifference in the image sizes used during training of OCR-Net vs size of the cropped logo images, original model being trained on english text.

 


We present the results of Levenshtein distance-based approach along with a vision-only encoder in Figure~\ref{fig:roc}, Table~\ref{tab:cropped_logo_results}, Table~\ref{tab:end_to_end}. In Table~\ref{tab:visual_backbone}, we present the comparison of Top-$k$ ($k=1$, 5 and 10) accuracy of our proposed encoder by varying visual encoders~\cite{krizhevsky2012imagenet, he2016deep} as backbones on the task of cropped logo identification on the QMUL-OpenLogo dataset. An encoder with our proposed fusion of both text and visual embeddings trained with the proposed loss formulation brings in the best from both modalities and induces better representative capabilities of the model, thereby resulting in noticeably superior performance over the baselines on unseen logo identification tasks at scale. 

\subsection{Qualitative Results}
\label{sec:qual}
We perform an extensive qualitative analysis of our framework on both cropped logo identification as well as end-to-end logo identification from natural scene images. A selection of visual results on cropped logo identification is shown in Figure~\ref{fig:visRes}; similarly, a selection of visual results on end-to-end logo identification on natural scene images is shown in Figure~\ref{fig:e2evisRes}.

\subsection{Limitations and Future scope}
\label{sec:limit}
We observe the following limitations of our work: (i) our proposed constrastive formulation of textual-visual features of logos is not tailored for time efficiency, (ii) we have used an off-the-shelf OCR-Net model to extract text from logos, which is trained and tested over English texts; hence, our model might suffer when logo images contain text from languages other than English, and (iii) the problem is far from solved when the scale is 100K in the task of large-scale open-set one-shot logo identification. We leave addressing these limitations as a future work.

\section{Conclusion}
Text within the logo has been underexplored for the task of \emph{Open-set One-shot Logo Identification}. Towards this end, we have presented a framework that fuses textual as well as visual features associated with the graphical design of logos and learns robust representation using a novel formulation of supervised contrastive learning. Our proposed method outperformed previous state-of-the-art methods under one-shot constraints. We have also introduced a large-scale logo dataset, \datafull{}, which has a potentially huge scope in benchmarking and evaluating large-scale open-set one-shot logo identification techniques. Furthermore, our exhaustive experiments have demonstrated that the representations learned by our framework are fairly robust compared to competent baselines on the task of large-scale open-set one-shot logo identification. We made our data and implementation publicly available for enabling future research. 

\begin{acks}
Abhirama S. Penamakuri is supported by Prime Minister Research Fellowship (PMRF), Ministry of Education, Government of India.
\end{acks}

\bibliographystyle{ACM-Reference-Format}
\bibliography{ICVGIP21-CameraReady-Template}


\begin{thebibliography}{44}


\ifx \showCODEN    \undefined \def \showCODEN     #1{\unskip}     \fi
\ifx \showDOI      \undefined \def \showDOI       #1{#1}\fi
\ifx \showISBNx    \undefined \def \showISBNx     #1{\unskip}     \fi
\ifx \showISBNxiii \undefined \def \showISBNxiii  #1{\unskip}     \fi
\ifx \showISSN     \undefined \def \showISSN      #1{\unskip}     \fi
\ifx \showLCCN     \undefined \def \showLCCN      #1{\unskip}     \fi
\ifx \shownote     \undefined \def \shownote      #1{#1}          \fi
\ifx \showarticletitle \undefined \def \showarticletitle #1{#1}   \fi
\ifx \showURL      \undefined \def \showURL       {\relax}        \fi
\providecommand\bibfield[2]{#2}
\providecommand\bibinfo[2]{#2}
\providecommand\natexlab[1]{#1}
\providecommand\showeprint[2][]{arXiv:#2}

\bibitem[\protect\citeauthoryear{Baek, Kim, Lee, Park, Han, Yun, Oh, and
  Lee}{Baek et~al\mbox{.}}{2019}]%
        {baek2019STRcomparisons}
\bibfield{author}{\bibinfo{person}{Jeonghun Baek}, \bibinfo{person}{Geewook
  Kim}, \bibinfo{person}{Junyeop Lee}, \bibinfo{person}{Sungrae Park},
  \bibinfo{person}{Dongyoon Han}, \bibinfo{person}{Sangdoo Yun},
  \bibinfo{person}{Seong~Joon Oh}, {and} \bibinfo{person}{Hwalsuk Lee}.}
  \bibinfo{year}{2019}\natexlab{}.
\newblock \showarticletitle{What Is Wrong With Scene Text Recognition Model
  Comparisons? Dataset and Model Analysis}. In
  \bibinfo{booktitle}{\emph{ICCV}}.
\newblock


\bibitem[\protect\citeauthoryear{Bastan, Wu, Cao, Kota, and Tek}{Bastan
  et~al\mbox{.}}{2019}]%
        {bastan2019large}
\bibfield{author}{\bibinfo{person}{Muhammet Bastan}, \bibinfo{person}{Hao-Yu
  Wu}, \bibinfo{person}{Tian Cao}, \bibinfo{person}{Bhargava Kota}, {and}
  \bibinfo{person}{Mehmet Tek}.} \bibinfo{year}{2019}\natexlab{}.
\newblock \showarticletitle{Large scale open-set deep logo detection}.
\newblock \bibinfo{journal}{\emph{arXiv preprint arXiv:1911.07440}}
  (\bibinfo{year}{2019}).
\newblock


\bibitem[\protect\citeauthoryear{Bhunia, Bhunia, Ghose, Das, Roy, and
  Pal}{Bhunia et~al\mbox{.}}{2019}]%
        {bhunia2019deep}
\bibfield{author}{\bibinfo{person}{Ayan~Kumar Bhunia},
  \bibinfo{person}{Ankan~Kumar Bhunia}, \bibinfo{person}{Shuvozit Ghose},
  \bibinfo{person}{Abhirup Das}, \bibinfo{person}{Partha~Pratim Roy}, {and}
  \bibinfo{person}{Umapada Pal}.} \bibinfo{year}{2019}\natexlab{}.
\newblock \showarticletitle{A deep one-shot network for query-based logo
  retrieval}.
\newblock \bibinfo{journal}{\emph{Pattern Recognition}}  \bibinfo{volume}{96}
  (\bibinfo{year}{2019}), \bibinfo{pages}{106965}.
\newblock


\bibitem[\protect\citeauthoryear{Bianco, Buzzelli, Mazzini, and
  Schettini}{Bianco et~al\mbox{.}}{2015}]%
        {bianco2015logo}
\bibfield{author}{\bibinfo{person}{Simone Bianco}, \bibinfo{person}{Marco
  Buzzelli}, \bibinfo{person}{Davide Mazzini}, {and} \bibinfo{person}{Raimondo
  Schettini}.} \bibinfo{year}{2015}\natexlab{}.
\newblock \showarticletitle{Logo recognition using cnn features}. In
  \bibinfo{booktitle}{\emph{International Conference on Image Analysis and
  Processing}}.
\newblock


\bibitem[\protect\citeauthoryear{Bianco, Buzzelli, Mazzini, and
  Schettini}{Bianco et~al\mbox{.}}{2017}]%
        {bianco2017deep}
\bibfield{author}{\bibinfo{person}{Simone Bianco}, \bibinfo{person}{Marco
  Buzzelli}, \bibinfo{person}{Davide Mazzini}, {and} \bibinfo{person}{Raimondo
  Schettini}.} \bibinfo{year}{2017}\natexlab{}.
\newblock \showarticletitle{Deep learning for logo recognition}.
\newblock \bibinfo{journal}{\emph{Neurocomputing}}  \bibinfo{volume}{245}
  (\bibinfo{year}{2017}), \bibinfo{pages}{23--30}.
\newblock


\bibitem[\protect\citeauthoryear{Chen, Kornblith, Norouzi, and Hinton}{Chen
  et~al\mbox{.}}{2020}]%
        {chen2020simple}
\bibfield{author}{\bibinfo{person}{Ting Chen}, \bibinfo{person}{Simon
  Kornblith}, \bibinfo{person}{Mohammad Norouzi}, {and}
  \bibinfo{person}{Geoffrey Hinton}.} \bibinfo{year}{2020}\natexlab{}.
\newblock \showarticletitle{A simple framework for contrastive learning of
  visual representations}. In \bibinfo{booktitle}{\emph{ICML}}.
\newblock


\bibitem[\protect\citeauthoryear{Chopra, Hadsell, and LeCun}{Chopra
  et~al\mbox{.}}{2005}]%
        {chopra2005learning}
\bibfield{author}{\bibinfo{person}{S. Chopra}, \bibinfo{person}{R. Hadsell},
  {and} \bibinfo{person}{Y. LeCun}.} \bibinfo{year}{2005}\natexlab{}.
\newblock \showarticletitle{Learning a similarity metric discriminatively, with
  application to face verification}. In \bibinfo{booktitle}{\emph{CVPR}}.
\newblock


\bibitem[\protect\citeauthoryear{Feh{\'e}rv{\'a}ri and
  Appalaraju}{Feh{\'e}rv{\'a}ri and Appalaraju}{2019}]%
        {fehervari2019scalable}
\bibfield{author}{\bibinfo{person}{Istv{\'a}n Feh{\'e}rv{\'a}ri} {and}
  \bibinfo{person}{Srikar Appalaraju}.} \bibinfo{year}{2019}\natexlab{}.
\newblock \showarticletitle{Scalable logo recognition using proxies}. In
  \bibinfo{booktitle}{\emph{WACV}}.
\newblock


\bibitem[\protect\citeauthoryear{Grill, Strub, Altch\'{e}, Tallec, Richemond,
  Buchatskaya, Doersch, Avila~Pires, Guo, Gheshlaghi~Azar, Piot, kavukcuoglu,
  Munos, and Valko}{Grill et~al\mbox{.}}{2020}]%
        {byol2020}
\bibfield{author}{\bibinfo{person}{Jean-Bastien Grill},
  \bibinfo{person}{Florian Strub}, \bibinfo{person}{Florent Altch\'{e}},
  \bibinfo{person}{Corentin Tallec}, \bibinfo{person}{Pierre Richemond},
  \bibinfo{person}{Elena Buchatskaya}, \bibinfo{person}{Carl Doersch},
  \bibinfo{person}{Bernardo Avila~Pires}, \bibinfo{person}{Zhaohan Guo},
  \bibinfo{person}{Mohammad Gheshlaghi~Azar}, \bibinfo{person}{Bilal Piot},
  \bibinfo{person}{koray kavukcuoglu}, \bibinfo{person}{Remi Munos}, {and}
  \bibinfo{person}{Michal Valko}.} \bibinfo{year}{2020}\natexlab{}.
\newblock \showarticletitle{Bootstrap Your Own Latent - A New Approach to
  Self-Supervised Learning}. In \bibinfo{booktitle}{\emph{NeurIPS}}.
\newblock


\bibitem[\protect\citeauthoryear{Gutmann and Hyv{\"a}rinen}{Gutmann and
  Hyv{\"a}rinen}{2010}]%
        {nce}
\bibfield{author}{\bibinfo{person}{Michael Gutmann} {and} \bibinfo{person}{Aapo
  Hyv{\"a}rinen}.} \bibinfo{year}{2010}\natexlab{}.
\newblock \showarticletitle{Noise-contrastive estimation: A new estimation
  principle for unnormalized statistical models}. In
  \bibinfo{booktitle}{\emph{Proceedings of the thirteenth international
  conference on artificial intelligence and statistics}}. JMLR Workshop and
  Conference Proceedings, \bibinfo{pages}{297--304}.
\newblock


\bibitem[\protect\citeauthoryear{Hadsell, Chopra, and LeCun}{Hadsell
  et~al\mbox{.}}{2006}]%
        {hadsell2006dimensionality}
\bibfield{author}{\bibinfo{person}{Raia Hadsell}, \bibinfo{person}{Sumit
  Chopra}, {and} \bibinfo{person}{Yann LeCun}.}
  \bibinfo{year}{2006}\natexlab{}.
\newblock \showarticletitle{Dimensionality reduction by learning an invariant
  mapping}. In \bibinfo{booktitle}{\emph{CVPR}}.
\newblock


\bibitem[\protect\citeauthoryear{He, Zhang, Ren, and Sun}{He
  et~al\mbox{.}}{2016}]%
        {he2016deep}
\bibfield{author}{\bibinfo{person}{Kaiming He}, \bibinfo{person}{Xiangyu
  Zhang}, \bibinfo{person}{Shaoqing Ren}, {and} \bibinfo{person}{Jian Sun}.}
  \bibinfo{year}{2016}\natexlab{}.
\newblock \showarticletitle{Deep residual learning for image recognition}. In
  \bibinfo{booktitle}{\emph{CVPR}}.
\newblock


\bibitem[\protect\citeauthoryear{Hoffer and Ailon}{Hoffer and Ailon}{2015}]%
        {hoffer2015deep}
\bibfield{author}{\bibinfo{person}{Elad Hoffer} {and} \bibinfo{person}{Nir
  Ailon}.} \bibinfo{year}{2015}\natexlab{}.
\newblock \showarticletitle{Deep metric learning using triplet network}. In
  \bibinfo{booktitle}{\emph{International workshop on similarity-based pattern
  recognition}}. Springer, \bibinfo{pages}{84--92}.
\newblock


\bibitem[\protect\citeauthoryear{Hoi, Wu, Liu, Wu, Wang, Xue, and Wu}{Hoi
  et~al\mbox{.}}{2015a}]%
        {hoi2015logo}
\bibfield{author}{\bibinfo{person}{Steven~CH Hoi}, \bibinfo{person}{Xiongwei
  Wu}, \bibinfo{person}{Hantang Liu}, \bibinfo{person}{Yue Wu},
  \bibinfo{person}{Huiqiong Wang}, \bibinfo{person}{Hui Xue}, {and}
  \bibinfo{person}{Qiang Wu}.} \bibinfo{year}{2015}\natexlab{a}.
\newblock \showarticletitle{Logo-net: Large-scale deep logo detection and brand
  recognition with deep region-based convolutional networks}.
\newblock \bibinfo{journal}{\emph{arXiv preprint arXiv:1511.02462}}
  (\bibinfo{year}{2015}).
\newblock


\bibitem[\protect\citeauthoryear{Hoi, Wu, Liu, Wu, Wang, Xue, and Wu}{Hoi
  et~al\mbox{.}}{2015b}]%
        {logonet}
\bibfield{author}{\bibinfo{person}{Steven~CH Hoi}, \bibinfo{person}{Xiongwei
  Wu}, \bibinfo{person}{Hantang Liu}, \bibinfo{person}{Yue Wu},
  \bibinfo{person}{Huiqiong Wang}, \bibinfo{person}{Hui Xue}, {and}
  \bibinfo{person}{Qiang Wu}.} \bibinfo{year}{2015}\natexlab{b}.
\newblock \showarticletitle{Logo-net: Large-scale deep logo detection and brand
  recognition with deep region-based convolutional networks}.
\newblock \bibinfo{journal}{\emph{arXiv preprint arXiv:1511.02462}}
  (\bibinfo{year}{2015}).
\newblock


\bibitem[\protect\citeauthoryear{Hou, Lin, Zhou, Qin, Jia, and Zheng}{Hou
  et~al\mbox{.}}{2017}]%
        {hou2017deep}
\bibfield{author}{\bibinfo{person}{Sujuan Hou}, \bibinfo{person}{Jianwei Lin},
  \bibinfo{person}{Shangbo Zhou}, \bibinfo{person}{Maoling Qin},
  \bibinfo{person}{Weikuan Jia}, {and} \bibinfo{person}{Yuanjie Zheng}.}
  \bibinfo{year}{2017}\natexlab{}.
\newblock \showarticletitle{Deep hierarchical representation from classifying
  logo-405}.
\newblock \bibinfo{journal}{\emph{Complexity}}  \bibinfo{volume}{2017}
  (\bibinfo{year}{2017}).
\newblock


\bibitem[\protect\citeauthoryear{Iandola, Shen, Gao, and Keutzer}{Iandola
  et~al\mbox{.}}{2015}]%
        {iandola2015deeplogo}
\bibfield{author}{\bibinfo{person}{Forrest~N Iandola}, \bibinfo{person}{Anting
  Shen}, \bibinfo{person}{Peter Gao}, {and} \bibinfo{person}{Kurt Keutzer}.}
  \bibinfo{year}{2015}\natexlab{}.
\newblock \showarticletitle{Deeplogo: Hitting logo recognition with the deep
  neural network hammer}.
\newblock \bibinfo{journal}{\emph{arXiv preprint arXiv:1510.02131}}
  (\bibinfo{year}{2015}).
\newblock


\bibitem[\protect\citeauthoryear{Jaiswal, Babu, Zadeh, Banerjee, and
  Makedon}{Jaiswal et~al\mbox{.}}{2021}]%
        {jaiswal2021survey}
\bibfield{author}{\bibinfo{person}{Ashish Jaiswal},
  \bibinfo{person}{Ashwin~Ramesh Babu}, \bibinfo{person}{Mohammad~Zaki Zadeh},
  \bibinfo{person}{Debapriya Banerjee}, {and} \bibinfo{person}{Fillia
  Makedon}.} \bibinfo{year}{2021}\natexlab{}.
\newblock \showarticletitle{A survey on contrastive self-supervised learning}.
\newblock \bibinfo{journal}{\emph{Technologies}} \bibinfo{volume}{9},
  \bibinfo{number}{1} (\bibinfo{year}{2021}), \bibinfo{pages}{2}.
\newblock


\bibitem[\protect\citeauthoryear{Jocher, Chaurasia, Stoken, Borovec,
  NanoCode012, Kwon, TaoXie, Fang, imyhxy, Michael, Lorna, V, Montes, Nadar,
  Laughing, tkianai, yxNONG, Skalski, Wang, Hogan, Fati, Mammana, AlexWang1900,
  Patel, Yiwei, You, Hajek, Diaconu, and Minh}{Jocher et~al\mbox{.}}{2022}]%
        {yolov5}
\bibfield{author}{\bibinfo{person}{Glenn Jocher}, \bibinfo{person}{Ayush
  Chaurasia}, \bibinfo{person}{Alex Stoken}, \bibinfo{person}{Jirka Borovec},
  \bibinfo{person}{NanoCode012}, \bibinfo{person}{Yonghye Kwon},
  \bibinfo{person}{TaoXie}, \bibinfo{person}{Jiacong Fang},
  \bibinfo{person}{imyhxy}, \bibinfo{person}{Kalen Michael},
  \bibinfo{person}{Lorna}, \bibinfo{person}{Abhiram V}, \bibinfo{person}{Diego
  Montes}, \bibinfo{person}{Jebastin Nadar}, \bibinfo{person}{Laughing},
  \bibinfo{person}{tkianai}, \bibinfo{person}{yxNONG}, \bibinfo{person}{Piotr
  Skalski}, \bibinfo{person}{Zhiqiang Wang}, \bibinfo{person}{Adam Hogan},
  \bibinfo{person}{Cristi Fati}, \bibinfo{person}{Lorenzo Mammana},
  \bibinfo{person}{AlexWang1900}, \bibinfo{person}{Deep Patel},
  \bibinfo{person}{Ding Yiwei}, \bibinfo{person}{Felix You},
  \bibinfo{person}{Jan Hajek}, \bibinfo{person}{Laurentiu Diaconu}, {and}
  \bibinfo{person}{Mai~Thanh Minh}.} \bibinfo{year}{2022}\natexlab{}.
\newblock \bibinfo{booktitle}{\emph{{ultralytics/yolov5: v6.1 - TensorRT,
  TensorFlow Edge TPU and OpenVINO Export and Inference}}}.
\newblock
\urldef\tempurl%
\url{https://doi.org/10.5281/zenodo.6222936}
\showDOI{\tempurl}


\bibitem[\protect\citeauthoryear{Joly and Buisson}{Joly and Buisson}{2009}]%
        {joly2009logo}
\bibfield{author}{\bibinfo{person}{Alexis Joly} {and} \bibinfo{person}{Olivier
  Buisson}.} \bibinfo{year}{2009}\natexlab{}.
\newblock \showarticletitle{Logo retrieval with a contrario visual query
  expansion}. In \bibinfo{booktitle}{\emph{ACM-MM}}.
\newblock


\bibitem[\protect\citeauthoryear{Kalantidis, Pueyo, Trevisiol, van Zwol, and
  Avrithis}{Kalantidis et~al\mbox{.}}{2011}]%
        {flickr27}
\bibfield{author}{\bibinfo{person}{Y. Kalantidis}, \bibinfo{person}{LG. Pueyo},
  \bibinfo{person}{M. Trevisiol}, \bibinfo{person}{R. van Zwol}, {and}
  \bibinfo{person}{Y. Avrithis}.} \bibinfo{year}{2011}\natexlab{}.
\newblock \showarticletitle{Scalable Triangulation-based Logo Recognition}. In
  \bibinfo{booktitle}{\emph{ICMR}}.
\newblock


\bibitem[\protect\citeauthoryear{Khosla, Teterwak, Wang, Sarna, Tian, Isola,
  Maschinot, Liu, and Krishnan}{Khosla et~al\mbox{.}}{2020}]%
        {khosla2020supervised}
\bibfield{author}{\bibinfo{person}{Prannay Khosla}, \bibinfo{person}{Piotr
  Teterwak}, \bibinfo{person}{Chen Wang}, \bibinfo{person}{Aaron Sarna},
  \bibinfo{person}{Yonglong Tian}, \bibinfo{person}{Phillip Isola},
  \bibinfo{person}{Aaron Maschinot}, \bibinfo{person}{Ce Liu}, {and}
  \bibinfo{person}{Dilip Krishnan}.} \bibinfo{year}{2020}\natexlab{}.
\newblock \showarticletitle{Supervised contrastive learning}.
\newblock \bibinfo{journal}{\emph{NeurIPS}} (\bibinfo{year}{2020}).
\newblock


\bibitem[\protect\citeauthoryear{Kim, Lee, Oh, and Kweon}{Kim
  et~al\mbox{.}}{2018}]%
        {quadnet}
\bibfield{author}{\bibinfo{person}{Junsik Kim}, \bibinfo{person}{Seokju Lee},
  \bibinfo{person}{Tae-Hyun Oh}, {and} \bibinfo{person}{In~So Kweon}.}
  \bibinfo{year}{2018}\natexlab{}.
\newblock \showarticletitle{Co-domain embedding using deep quadruplet networks
  for unseen traffic sign recognition}. In \bibinfo{booktitle}{\emph{AAAI}}.
\newblock


\bibitem[\protect\citeauthoryear{Kim, Oh, Lee, Pan, and Kweon}{Kim
  et~al\mbox{.}}{2019}]%
        {vpe}
\bibfield{author}{\bibinfo{person}{Junsik Kim}, \bibinfo{person}{Tae-Hyun Oh},
  \bibinfo{person}{Seokju Lee}, \bibinfo{person}{Fei Pan}, {and}
  \bibinfo{person}{In~So Kweon}.} \bibinfo{year}{2019}\natexlab{}.
\newblock \showarticletitle{Variational prototyping-encoder: One-shot learning
  with prototypical images}. In \bibinfo{booktitle}{\emph{CVPR}}.
\newblock


\bibitem[\protect\citeauthoryear{Koch, Zemel, Salakhutdinov,
  et~al\mbox{.}}{Koch et~al\mbox{.}}{2015}]%
        {koch2015siamese}
\bibfield{author}{\bibinfo{person}{Gregory Koch}, \bibinfo{person}{Richard
  Zemel}, \bibinfo{person}{Ruslan Salakhutdinov}, {et~al\mbox{.}}}
  \bibinfo{year}{2015}\natexlab{}.
\newblock \showarticletitle{Siamese neural networks for one-shot image
  recognition}. In \bibinfo{booktitle}{\emph{ICML deep learning workshop}}.
\newblock


\bibitem[\protect\citeauthoryear{Krizhevsky, Sutskever, and Hinton}{Krizhevsky
  et~al\mbox{.}}{2012}]%
        {krizhevsky2012imagenet}
\bibfield{author}{\bibinfo{person}{Alex Krizhevsky}, \bibinfo{person}{Ilya
  Sutskever}, {and} \bibinfo{person}{Geoffrey~E Hinton}.}
  \bibinfo{year}{2012}\natexlab{}.
\newblock \showarticletitle{Imagenet classification with deep convolutional
  neural networks}.
\newblock \bibinfo{journal}{\emph{NeurIPS}} (\bibinfo{year}{2012}).
\newblock


\bibitem[\protect\citeauthoryear{Li, Feh{\'e}rv{\'a}ri, Zhao, Macedo, and
  Appalaraju}{Li et~al\mbox{.}}{2022}]%
        {li2022seetek}
\bibfield{author}{\bibinfo{person}{Chenge Li}, \bibinfo{person}{Istv{\'a}n
  Feh{\'e}rv{\'a}ri}, \bibinfo{person}{Xiaonan Zhao}, \bibinfo{person}{Ives
  Macedo}, {and} \bibinfo{person}{Srikar Appalaraju}.}
  \bibinfo{year}{2022}\natexlab{}.
\newblock \showarticletitle{SeeTek: Very Large-Scale Open-set Logo Recognition
  with Text-Aware Metric Learning}. In \bibinfo{booktitle}{\emph{WACV}}.
\newblock


\bibitem[\protect\citeauthoryear{Neumann, Samet, and Soffer}{Neumann
  et~al\mbox{.}}{2002}]%
        {belgalogos}
\bibfield{author}{\bibinfo{person}{Jan Neumann}, \bibinfo{person}{Hanan Samet},
  {and} \bibinfo{person}{Aya Soffer}.} \bibinfo{year}{2002}\natexlab{}.
\newblock \showarticletitle{Integration of local and global shape analysis for
  logo classification}.
\newblock \bibinfo{journal}{\emph{Pattern recognition letters}}
  \bibinfo{volume}{23}, \bibinfo{number}{12} (\bibinfo{year}{2002}),
  \bibinfo{pages}{1449--1457}.
\newblock


\bibitem[\protect\citeauthoryear{Romberg and Lienhart}{Romberg and
  Lienhart}{2013}]%
        {romberg2013bundle}
\bibfield{author}{\bibinfo{person}{Stefan Romberg} {and}
  \bibinfo{person}{Rainer Lienhart}.} \bibinfo{year}{2013}\natexlab{}.
\newblock \showarticletitle{Bundle min-hashing for logo recognition}. In
  \bibinfo{booktitle}{\emph{ICMR}}.
\newblock


\bibitem[\protect\citeauthoryear{Romberg, Pueyo, Lienhart, and
  Van~Zwol}{Romberg et~al\mbox{.}}{2011}]%
        {romberg2011scalable}
\bibfield{author}{\bibinfo{person}{Stefan Romberg},
  \bibinfo{person}{Lluis~Garcia Pueyo}, \bibinfo{person}{Rainer Lienhart},
  {and} \bibinfo{person}{Roelof Van~Zwol}.} \bibinfo{year}{2011}\natexlab{}.
\newblock \showarticletitle{Scalable logo recognition in real-world images}. In
  \bibinfo{booktitle}{\emph{ICMR}}.
\newblock


\bibitem[\protect\citeauthoryear{Shi, Bai, and Yao}{Shi et~al\mbox{.}}{2017}]%
        {Shi2017AnET}
\bibfield{author}{\bibinfo{person}{Baoguang Shi}, \bibinfo{person}{Xiang Bai},
  {and} \bibinfo{person}{Cong Yao}.} \bibinfo{year}{2017}\natexlab{}.
\newblock \showarticletitle{An End-to-End Trainable Neural Network for
  Image-Based Sequence Recognition and Its Application to Scene Text
  Recognition}.
\newblock \bibinfo{journal}{\emph{IEEE TPAMI}}  \bibinfo{volume}{39}
  (\bibinfo{year}{2017}), \bibinfo{pages}{2298--2304}.
\newblock


\bibitem[\protect\citeauthoryear{Sohn}{Sohn}{2016}]%
        {npairloss}
\bibfield{author}{\bibinfo{person}{Kihyuk Sohn}.}
  \bibinfo{year}{2016}\natexlab{}.
\newblock \showarticletitle{Improved Deep Metric Learning with Multi-class
  N-pair Loss Objective}. In \bibinfo{booktitle}{\emph{NeurIPS}},
  \bibfield{editor}{\bibinfo{person}{Daniel~D. Lee}, \bibinfo{person}{Masashi
  Sugiyama}, \bibinfo{person}{Ulrike von Luxburg}, \bibinfo{person}{Isabelle
  Guyon}, {and} \bibinfo{person}{Roman Garnett}} (Eds.).
\newblock


\bibitem[\protect\citeauthoryear{Su, Gong, and Zhu}{Su et~al\mbox{.}}{2017a}]%
        {su2017weblogo}
\bibfield{author}{\bibinfo{person}{Hang Su}, \bibinfo{person}{Shaogang Gong},
  {and} \bibinfo{person}{Xiatian Zhu}.} \bibinfo{year}{2017}\natexlab{a}.
\newblock \showarticletitle{Weblogo-2m: Scalable logo detection by deep
  learning from the web}. In \bibinfo{booktitle}{\emph{CVPRW}}.
\newblock


\bibitem[\protect\citeauthoryear{Su, Zhu, and Gong}{Su et~al\mbox{.}}{2017b}]%
        {su2017deep}
\bibfield{author}{\bibinfo{person}{Hang Su}, \bibinfo{person}{Xiatian Zhu},
  {and} \bibinfo{person}{Shaogang Gong}.} \bibinfo{year}{2017}\natexlab{b}.
\newblock \showarticletitle{Deep learning logo detection with data expansion by
  synthesising context}. In \bibinfo{booktitle}{\emph{WACV}}.
\newblock


\bibitem[\protect\citeauthoryear{Su, Zhu, and Gong}{Su et~al\mbox{.}}{2018}]%
        {su2018open}
\bibfield{author}{\bibinfo{person}{Hang Su}, \bibinfo{person}{Xiatian Zhu},
  {and} \bibinfo{person}{Shaogang Gong}.} \bibinfo{year}{2018}\natexlab{}.
\newblock \showarticletitle{Open Logo Detection Challenge}. In
  \bibinfo{booktitle}{\emph{BMVC}}.
\newblock


\bibitem[\protect\citeauthoryear{Tian, Krishnan, and Isola}{Tian
  et~al\mbox{.}}{2020}]%
        {tian2020contrastive}
\bibfield{author}{\bibinfo{person}{Yonglong Tian}, \bibinfo{person}{Dilip
  Krishnan}, {and} \bibinfo{person}{Phillip Isola}.}
  \bibinfo{year}{2020}\natexlab{}.
\newblock \showarticletitle{Contrastive Multiview Coding}. In
  \bibinfo{booktitle}{\emph{ECCV}}.
\newblock


\bibitem[\protect\citeauthoryear{T{\"u}zk{\"o}, Herrmann, Manger, and
  Beyerer}{T{\"u}zk{\"o} et~al\mbox{.}}{2017}]%
        {tuzko2017open}
\bibfield{author}{\bibinfo{person}{Andras T{\"u}zk{\"o}},
  \bibinfo{person}{Christian Herrmann}, \bibinfo{person}{Daniel Manger}, {and}
  \bibinfo{person}{J{\"u}rgen Beyerer}.} \bibinfo{year}{2017}\natexlab{}.
\newblock \showarticletitle{Open set logo detection and retrieval}.
\newblock \bibinfo{journal}{\emph{arXiv preprint arXiv:1710.10891}}
  (\bibinfo{year}{2017}).
\newblock


\bibitem[\protect\citeauthoryear{Vargas, Zhang, and Izquierdo}{Vargas
  et~al\mbox{.}}{2020}]%
        {vargas2020one}
\bibfield{author}{\bibinfo{person}{Camilo Vargas}, \bibinfo{person}{Qianni
  Zhang}, {and} \bibinfo{person}{Ebroul Izquierdo}.}
  \bibinfo{year}{2020}\natexlab{}.
\newblock \showarticletitle{One shot logo recognition based on siamese neural
  networks}. In \bibinfo{booktitle}{\emph{ICMR}}.
\newblock


\bibitem[\protect\citeauthoryear{Vinyals, Blundell, Lillicrap, Wierstra,
  et~al\mbox{.}}{Vinyals et~al\mbox{.}}{2016}]%
        {matchnet}
\bibfield{author}{\bibinfo{person}{Oriol Vinyals}, \bibinfo{person}{Charles
  Blundell}, \bibinfo{person}{Timothy Lillicrap}, \bibinfo{person}{Daan
  Wierstra}, {et~al\mbox{.}}} \bibinfo{year}{2016}\natexlab{}.
\newblock \showarticletitle{Matching networks for one shot learning}.
\newblock \bibinfo{journal}{\emph{NeurIPS}} (\bibinfo{year}{2016}).
\newblock


\bibitem[\protect\citeauthoryear{Vrande{\v{c}}i{\'c} and
  Kr{\"o}tzsch}{Vrande{\v{c}}i{\'c} and Kr{\"o}tzsch}{2014}]%
        {wikidata}
\bibfield{author}{\bibinfo{person}{Denny Vrande{\v{c}}i{\'c}} {and}
  \bibinfo{person}{Markus Kr{\"o}tzsch}.} \bibinfo{year}{2014}\natexlab{}.
\newblock \showarticletitle{Wikidata: a free collaborative knowledgebase}.
\newblock \bibinfo{journal}{\emph{Commun. ACM}} \bibinfo{volume}{57},
  \bibinfo{number}{10} (\bibinfo{year}{2014}), \bibinfo{pages}{78--85}.
\newblock


\bibitem[\protect\citeauthoryear{Wang, Min, Hou, Ma, Zheng, and Jiang}{Wang
  et~al\mbox{.}}{2022}]%
        {wang2022logodet}
\bibfield{author}{\bibinfo{person}{Jing Wang}, \bibinfo{person}{Weiqing Min},
  \bibinfo{person}{Sujuan Hou}, \bibinfo{person}{Shengnan Ma},
  \bibinfo{person}{Yuanjie Zheng}, {and} \bibinfo{person}{Shuqiang Jiang}.}
  \bibinfo{year}{2022}\natexlab{}.
\newblock \showarticletitle{LogoDet-3K: A Large-Scale Image Dataset for Logo
  Detection}.
\newblock \bibinfo{journal}{\emph{ACM Transactions on Multimedia Computing,
  Communications, and Applications (TOMM)}} \bibinfo{volume}{18},
  \bibinfo{number}{1} (\bibinfo{year}{2022}), \bibinfo{pages}{1--19}.
\newblock


\bibitem[\protect\citeauthoryear{Wang, Min, Hou, Ma, Zheng, Wang, and
  Jiang}{Wang et~al\mbox{.}}{2020}]%
        {logo2k}
\bibfield{author}{\bibinfo{person}{Jing Wang}, \bibinfo{person}{Weiqing Min},
  \bibinfo{person}{Sujuan Hou}, \bibinfo{person}{Shengnan Ma},
  \bibinfo{person}{Yuanjie Zheng}, \bibinfo{person}{Haishuai Wang}, {and}
  \bibinfo{person}{Shuqiang Jiang}.} \bibinfo{year}{2020}\natexlab{}.
\newblock \showarticletitle{Logo-2K+: A large-scale logo dataset for scalable
  logo classification}. In \bibinfo{booktitle}{\emph{AAAI}}.
\newblock


\bibitem[\protect\citeauthoryear{Xiao, Madapana, and Wachs}{Xiao
  et~al\mbox{.}}{2021}]%
        {vpe++}
\bibfield{author}{\bibinfo{person}{Chenxi Xiao}, \bibinfo{person}{Naveen
  Madapana}, {and} \bibinfo{person}{Juan Wachs}.}
  \bibinfo{year}{2021}\natexlab{}.
\newblock \showarticletitle{One-Shot Image Recognition Using Prototypical
  Encoders with Reduced Hubness}. In \bibinfo{booktitle}{\emph{WACV}}.
\newblock


\bibitem[\protect\citeauthoryear{Zbontar, Jing, Misra, LeCun, and Deny}{Zbontar
  et~al\mbox{.}}{2021}]%
        {pmlr-v139-zbontar21a}
\bibfield{author}{\bibinfo{person}{Jure Zbontar}, \bibinfo{person}{Li Jing},
  \bibinfo{person}{Ishan Misra}, \bibinfo{person}{Yann LeCun}, {and}
  \bibinfo{person}{Stephane Deny}.} \bibinfo{year}{2021}\natexlab{}.
\newblock \showarticletitle{Barlow Twins: Self-Supervised Learning via
  Redundancy Reduction}. In \bibinfo{booktitle}{\emph{ICML}}.
\newblock


\end{thebibliography}

\end{document}